\documentclass[10pt,twocolumn,letterpaper]{article}

\usepackage{iccv}
\usepackage{times}
\usepackage{epsfig}
\usepackage{graphicx}
\usepackage{amsmath}
\usepackage{amssymb}
\usepackage{subfigure}

\usepackage{comment}

\usepackage[pagebackref=true,breaklinks=true,letterpaper=true,colorlinks,bookmarks=false]{hyperref}
\iccvfinalcopy 

\def\iccvPaperID{208} 

\ificcvfinal\fi
\begin{document}

\title{Learning Temporal Embeddings for Complex Video Analysis}

\author{Vignesh Ramanathan$^{1}$, Kevin Tang$^{1}$,
  Greg Mori$^{2}$ and Li Fei-Fei$^{1}$\\
  \\
  $^{1}$Stanford University, $^{2}$Simon Fraser University \\
  {\tt\small \{vigneshr, kdtang\}@cs.stanford.edu,
  mori@cs.sfu.ca, feifeili@cs.stanford.edu}
}
\maketitle

\begin{abstract}
In this paper, we propose to learn temporal embeddings of video frames for complex video analysis. Large quantities of unlabeled video data can be easily obtained from the Internet. These videos possess the implicit weak label that they are sequences of temporally and semantically coherent images. We leverage this information to learn temporal embeddings for video frames by associating frames with the temporal context that they appear in. To do this, we propose a scheme for incorporating temporal context based on past and future frames in videos, and compare this to other contextual representations. In addition, we show how data augmentation using multi-resolution samples and hard negatives helps to significantly improve the quality of the learned embeddings. We evaluate various design decisions for learning temporal embeddings, and show that our embeddings can improve performance for multiple video tasks such as retrieval, classification, and temporal order recovery in unconstrained Internet video. 
\end{abstract}

\section{Introduction}
Video data is plentiful and a ready source of information -- what can
we glean from watching massive quantities of videos?  At a fine
granularity, consecutive video frames are visually similar due to temporal
coherence.  At a coarser level, consecutive video frames are visually
distinct but semantically coherent.

Learning from this semantic coherence present in video at the coarser-level is the main focus of this
paper.  Purely from unlabeled video data, we aim to learn embeddings
for video frames that capture semantic similarity by using the temporal structure in videos.
The prospect of learning a generic embedding for video
frames holds promise for a variety of applications ranging
from generic retrieval and similarity measurement, video recommendation,
to automatic content creation such as summarization or collaging.
In this paper, we demonstrate the utility of our video frame
embeddings for several tasks such as video retrieval, classification
and temporal order recovery.

\begin{figure}[t!]
\centering
   \includegraphics[width=0.8\linewidth]{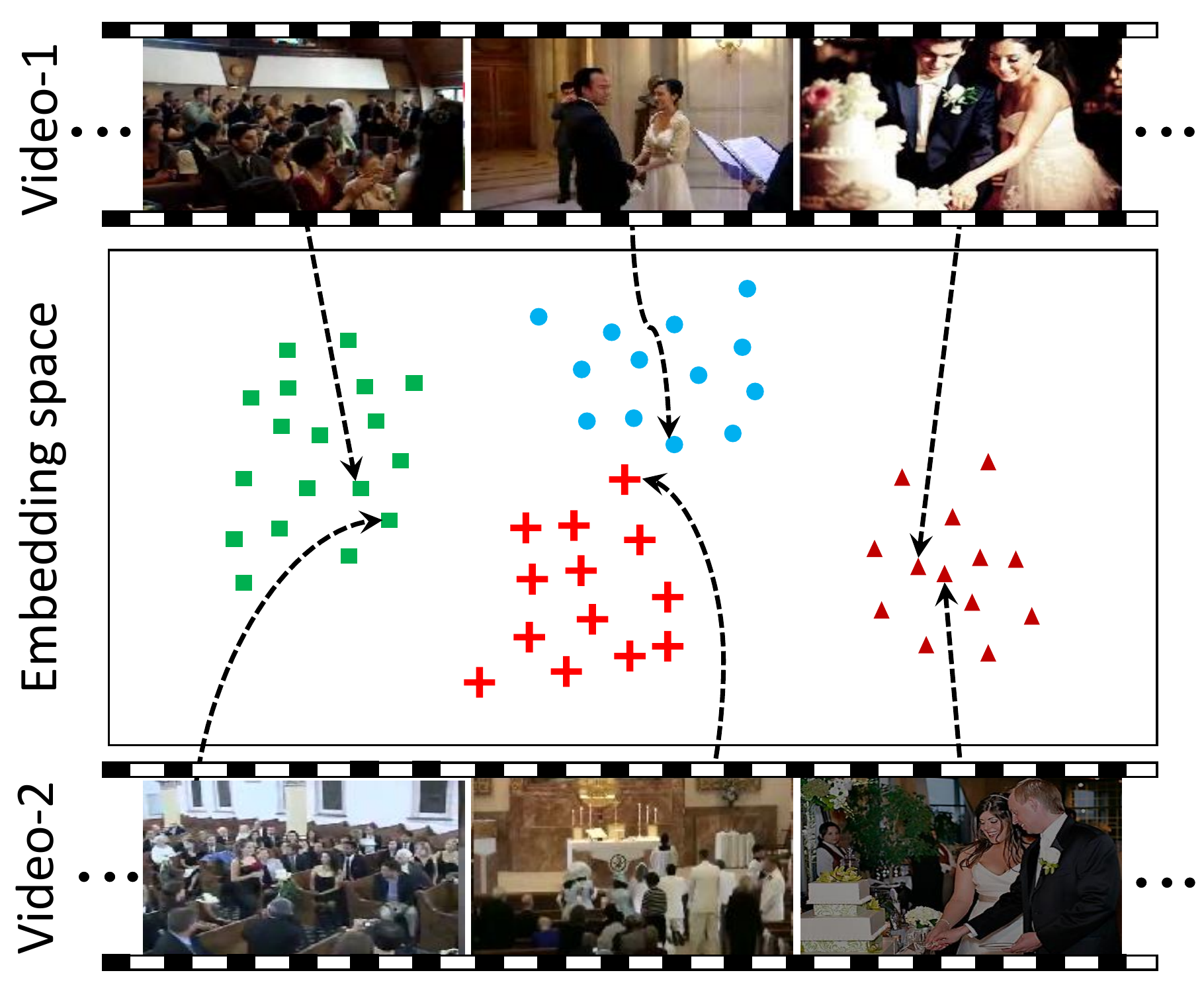}
   \caption{The temporal context of a video frame
     is crucial in determining its true semantic meaning.
 For instance, consider the above example
 where the embeddings of different semantic classes
 are shown in different colors.
 The middle frame from the two wedding videos correspond to visually
 dissimilar classes of ``church ceremony" and ``court ceremony".
 However, by observing the similarity in their temporal contexts
 we expect them to be semantically closer.
 Our work leverages such powerful temporal
 context to learn semantically rich embeddings.
}
\label{fig:pull_figure}
\end{figure}

The idea of leveraging sequential data to learn embeddings in an unsupervised
fashion is well explored in the Natural Language Processing (NLP) community.
In particular, distributed word vector representations such as word2vec \cite{Mikolov_2013}
have the unique ability to encode \emph{regularities and patterns} surrounding words,
using large amounts of unlabeled data.  In the embedding space, this
brings together words that may be very different, but which share
similar contexts in different sentences.  This is a desirable property
we would like to extend to video frames as well as
shown in Fig.~\ref{fig:pull_figure}. We would like to
have a representation for frames which captures the semantic context
around the frame beyond the visual similarity obtained from temporal
coherence.

However, the task of embedding frames poses multiple challenges
specific to the video domain:
1. Unlike words, the set of frames across all videos is not discrete
and quantizing the frames leads to a loss in information;
2. Temporally proximal frames within the same video
are often visually similar and might not provide useful
contextual information; 3. The correct representation
of context surrounding a frame is not obvious in videos.
The main contribution of our work is to
propose a new ranking loss based embedding framework, along
with a contextual representation specific to videos.
We also develop a well engineered
data augmentation strategy to promote visual
diversity among the context frames used for embedding.

We evaluate our learned embeddings
on the standard tasks of video event retrieval and classification 
on the TRECVID MED 2011 \cite{MED11} dataset, and compare
to several recently published spatial and temporal video representations \cite{Izadinia_ECCV12,Simonyan_2014}.
Aside from semantic similarity,
the learned embeddings capture valuable information
in terms of the temporal context shared between frames.
Hence, we also evaluate our embeddings on two related tasks:
1. temporal frame retrieval, and 2. temporal order recovery in videos.
Our embeddings improve performance on all tasks,
and serves as a powerful representation for video frames.

\section{Related Work}

\noindent \textbf{Video features.}
Standard tasks in video such as classification and retrieval require
a well engineered feature representation, with many proposed in the literature \cite{Dalal_ECCV06,Jain_CVPR13,Jiang_ECCV12,Laptev_CVPR08,
Niebels_ECCV10,Oh_MVA14,Oneata_ICCV13,Peng_ECCV14,Sadanand_CVPR12,Wang_BMVC09,Wang_CVPR11}.
Deep network features learned
from spatial data \cite{Ji_PAMI13,Karpathy_CVPR14,Simonyan_2014} and temporal flow \cite{Simonyan_2014}
have also shown comparable results.
However, recent works in complex event recognition \cite{Xu_2015,Zha_2015}
have shown that spatial Convolutional Neural Network (CNN) features learned from ImageNet \cite{Deng_CVPR09} without fine-tuning on video, 
accompanied by suitable pooling and encoding strategies achieves
state-of-the-art performance.
In contrast to these methods which either propose
handcrafted features or learn feature representations with a fully supervised
objective from images or videos, we try to learn an embedding in an unsupervised fashion.
Moreover, our learned features can be extended to other tasks beyond
classification and retrieval.

There are several works which improve
complex event recognition by combining multiple feature
modalities \cite{Jiang_ACM12,Natarajan_CVPR12,Tamrakar_CVPR12}.
Another related line of work is the use of
sub-events defined manually \cite{Izadinia_ECCV12},
or clustered from data \cite{Lai_ECCV14} to
improve recognition.
Similarly, Yang et al. used
low dimensional features from deep belief nets
and sparse coding \cite{Yang_ECCV12}.
While these methods are targeted towards building
features specifically for classification in limited settings,
we propose a generic video frame representation which
can capture semantic and temporal structure in videos. 



\noindent \textbf{Unsupervised learning in videos.}
Learning features with unsupervised objectives has been a challenging task
in the image and video domain \cite{Huang_ICCV07,Le_CVPR11,Taylor_ECCV10}.
Notably, \cite{Le_CVPR11} develops an Independent Subspace Analysis (ISA) model 
for feature learning using unlabeled video.
Recent work from \cite{Goroshin_2015} also
hints at a similar approach to exploit the slowness prior
in videos.
Also, recent attempts extend such autoencoder techniques for next frame prediction in videos
\cite{Ranzato_2014,Srivastava_2015}.
These methods try to capitalize on the temporal continuity in videos
to learn an LSTM \cite{Zaremba_2014} representation for frame prediction.
In contrast to these methods which aim to provide a unified representation for a
complete temporal sequence, our work provides a simple yet powerful representation
for independent video frames and images.

\noindent \textbf{Embedding models.}
The idea of embedding words to a dense lower dimension vector space has been
prevalent in the NLP community.
The word2vec model \cite{Mikolov_2013} tries to
learn embeddings such that words with similar contexts in sentences are
closer to each other. 
A related idea in computer vision is
the embedding of text in the semantic visual
space attempted by \cite{Frome_NIPS13,Kiros_2014}
based on large image datasets labeled with
captions or class names.
While these methods focus on different scenarios for embedding text, the aim of our work
is to generate an embedding for video frames.

\section {Our Method}

Given a large collection of unlabeled videos, our goal is to leverage their temporal structure to learn
an effective embedding for video frames.
We wish to learn an embedding such that the \emph{context} frames surrounding each \emph{target} frame can determine the representation of the \emph{target} frame, similar to the intuition from word2vec \cite{Mikolov_2013}.
For example, in Fig.~\ref{fig:pull_figure}, \emph{context} such as ``crowd" and ``cutting the cake" provides valuable information about the \emph{target} ``ceremony" frames that occur in between.
This idea is fundamental to our embedding objective and helps in capturing semantic and temporal
interactions in video.

While the idea of representing frames by embeddings is lucrative, the extension 
from language to visual data is not straightforward. Unlike language
we do not have a natural, discrete vocabulary of words.
This prevents us from using a softmax objective as in the case of
word2vec \cite{Mikolov_2013}.
Further, consecutive frames in videos
often share visual similarity due to temporal coherence. Hence,
a naive extension of \cite{Mikolov_2013} does not lead to good vector representations of frames.

To overcome the problem of lack of discrete words,
we use a ranking loss which explicitly
compares multiple pairs of frames across all videos in the dataset.
This ensures that the \emph{context} in a video scores
the \emph{target} frame higher than others in the dataset.
We also handle the problem of visually similar frames in temporally
smooth videos through a carefully designed sampling mechanism.
We obtain context frames by sampling the video at multiple
temporal scales, and choosing hard negatives from the same video.

\subsection{Embedding objective}
We are given a collection of videos $\mathcal{V}$,
where each video $v \in \mathcal{V}$ is a sequence of
frames $\{s_{v1}, \dots , s_{vn}\}$.
We wish to obtain an
embedding $f_{vj}$ for each frame $s_{vj}$. 
Let $f_{vj} = f(s_{vj}; W_e)$ be the temporal embedding function
which maps the frame $s_{vj}$ to a vector.
The model embedding parameters are given by $W_e$,
and will be learned by our method.
We embed the frames such that the \emph{context} frames around the \emph{target} frame 
predict the \emph{target} frame better than other frames.
The model is learned by minimizing the sum of objectives
across all videos.
Our embedding loss objective is shown below:

\begin{eqnarray}
  J(W_e) = \sum_{v \in \mathcal{V}} \sum \limits_{ \substack{ s_{vj} \in v \\ s_{-} \neq s_{vj}} } \max \left(0,
  1 - \left( f_{vj} - f_{-} \right) \cdot h_{vj} \right),
\end{eqnarray}where $f_{-}$ is the embedding of
a negative frame $s_{-}$, and
the context surrounding the frame $s_{vj}$ is represented
by the vector $h_{vj}$. Note that unlike the word vector
embedding models in word2vec \cite{Mikolov_2013}, we do not use
an additional linear layer for softmax prediction on top of the context vector.

\begin{figure}[t!]
\centering
   \includegraphics[width=1\linewidth]{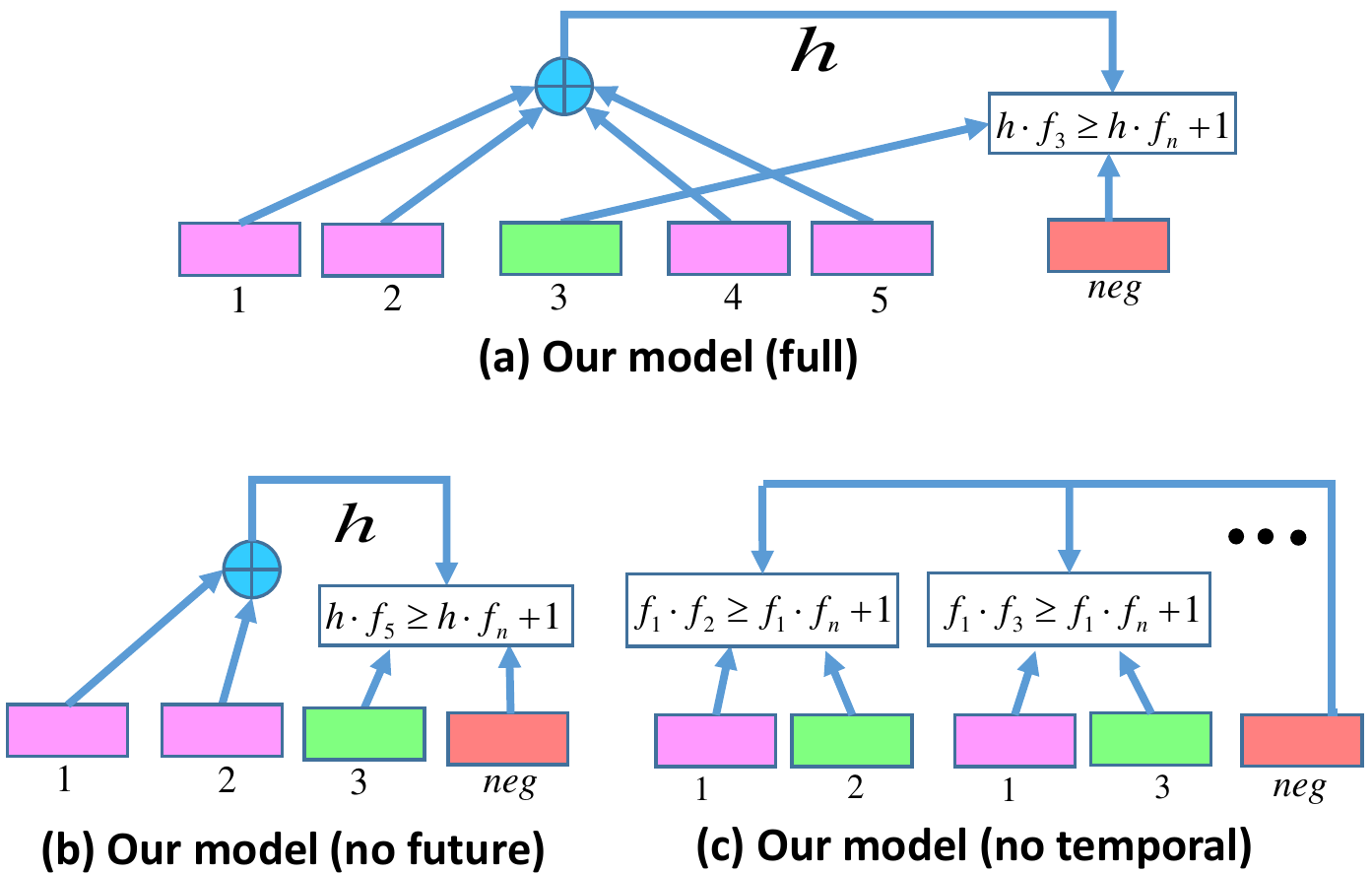}
   \caption{Visualizations of the temporal context of frames used in: 
   (a) our model (full),
   (b) our model (no future),
   and (c) our model (no temporal). Green boxes denote target frames, magenta boxes denote contextual frames, and red boxes denote negative frames.}
   \label{fig:system_figure_1}
\end{figure}

\subsection{Context representation}

As we verify later in the experiments, the choice of
context is crucial to learning good embeddings.
A video frame at any time instant is semantically
correlated with both past and future frames in the video.
Hence, a natural choice for context representation would
involve a window of frames centered around the \emph{target} frame,
similar to the skip-gram idea used in word2vec \cite{Mikolov_2013}.
Along these lines, we propose a 
context representation given by the average of the frame
embeddings around the \emph{target} frame.
Our context vector $h_{vj}$ for a frame $s_{vj}$
is:

\begin{eqnarray}
  h_{vj} & = & \frac{1}{2T} \sum_{t=1}^{T} f_{vj+t} + f_{vj-t},
\end{eqnarray} where $T$ is the window size, and $f_{vj}$ is the embedding of the frame 
$s_{vj}$. This embedding
model is shown in Fig.~\ref{fig:system_figure_1}(a).
For negatives,
we use frames from other videos as well as frames from the same video
which are outside the temporal window,
as explained in Sec.~\ref{sec:negative_multi_res}.

Two important characteristics of this context representation
is that it 1. makes use of the temporal order in which frames
occur and 2. considers contextual evidence from both past and future.
In order to examine their effect on the quality of the learned embedding, we also consider
two weaker variants of the context representation below.

\noindent \textbf{Our model (no future).}
This one-sided contextual representation tries to predict the
embedding of a frame in a video only based on
the embeddings of frames from the past
as shown in Fig.~\ref{fig:system_figure_1}(b).
For a frame $s_{vj}$, the context
$h_{vj}^{nofuture}$
is given by:

\begin{eqnarray}
  h_{vj}^{nofuture} & = & \frac{1}{T} \sum_{t=1}^{T} f_{vj-t},
\end{eqnarray} where $T$ is the window size.


\noindent \textbf{Our model (no temporal).}
An even weaker variant of context representation is simple co-occurrence
without temporal information.
We also explore a contextual representation
which completely neglects the temporal ordering of frames
and treats a video as a bag of frames.
The context $h_{vj}^{notemp}$ for a target frame
$s_{vj}$ is sampled from the
embeddings corresponding to all other frames
in the same video:

\begin{eqnarray}
  h_{vj}^{notemp} \in \{ f_{vk} \;\; | \;\; k \neq j \}.
\end{eqnarray} 
This contextual representation is visualized in Fig.~\ref{fig:system_figure_1}(c).

\subsection{Embedding function}

In the previous sections, we introduced a model for representing context,
and now move on to discuss the embedding function $f(s_{ij}; W_e) $.
In practice, the embedding function can be a CNN built from the frame pixels, or any underlying image or video representation.
However, following the recent success of ImageNet trained CNN features for complex event videos \cite{Xu_2015,Zha_2015}, we choose to learn an embedding on top of the fully connected
fc6 layer feature representation
obtained by passing the frame through a standard
CNN \cite{Krizhevsky_NIPS12} architecture.
We use a simple model with a fully connected layer followed by a rectified linear unit (ReLU)
and local response normalization (LRN) layer, with dropout regularization.
In this architecture, the learned model parameters $W_e$ correspond to the weights and bias of our affine layer.

\subsection{Data augmentation}
\label{sec:negative_multi_res}
We found that a careful strategy for sampling context frames and negatives is important 
to learning high quality embeddings in our
models. This helps both in handling the problem of temporal
smoothness and prevents the model from
overfitting to less interesting video-specific properties.


\begin{figure}[t]
   \includegraphics[width=1\linewidth]{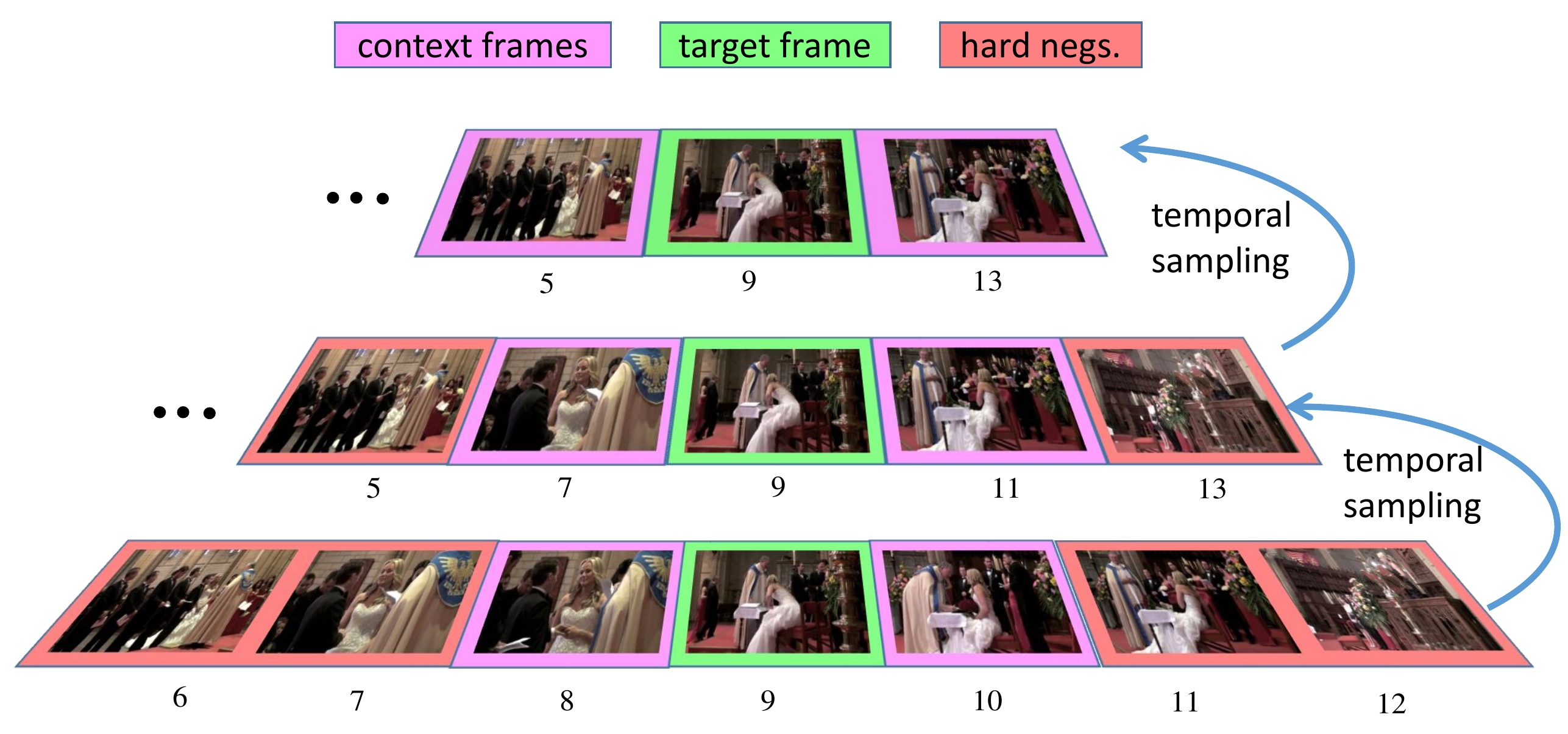}
   \caption{Multi-resolution sampling and hard negatives used in our full context model ($T$ = 1).
   For a target frame (green),
we sample context frames (magenta) at varying resolutions,
as shown by the rows in this figure.
We take hard negatives as examples in the same
video that fall outside the context window (red).
}
\label{fig:system_figure_2}
\vspace{-12pt}
\end{figure}


\noindent \textbf{Multi-resolution sampling.}
Complex events progress at different paces within different videos.
Densely sampling frames in slowly changing videos can
lead to context windows comprised of frames
that are visually very similar to the target frame.
On the other hand, a sparse
sampling of fast videos could lead to 
context windows only composed of
disjoint frames from unrelated parts of the video.
We overcome these problems through multi-resolution
sampling as shown in Fig.~\ref{fig:system_figure_2}.
For every target frame, we sample context frames from multiple
temporal resolutions. This ensures a good trade-off between
visual variety and semantic relatedness
in the context windows.

\noindent \textbf{Hard negatives.}
The context frames, as well as the target to be
scored are chosen from the same video. This
causes the model to cluster frames
from the same video based
on less interesting video-specific properties
such as lighting, camera characteristics
and background, without learning anything semantically
meaningful. We avoid such problems by choosing hard
negatives from within the same video as well.
Empirically, this improves performance
for all tasks.
The negatives are chosen from outside the range
of the context window
within a video
as depicted in Fig.~\ref{fig:system_figure_2}.

\subsection{Implementation details}
The context window size was set to $T=2$, and the embedding dimension to $4096$.
The learning rate was set to $0.01$
and gradually annealed in steps of $5000$. The training
is typically completed within a day on 1 GPU with Caffe \cite{Jia_2014} for
a dataset of approximately $40000$ videos.
All videos were first down-sampled to $0.2$ fps before training.
The embedding code as
well as the learned models and video embeddings will be made
publicly available upon publication.

\section {Experimental Setup}

Our embeddings are aimed at capturing semantic and temporal interactions
within complex events in a video, and thus we require a generic set of videos with
a good variety of actions and sub-events within each video.
Most standard datasets such as UCF-101 \cite{UCF_101} and Sport-1M \cite{Karpathy_CVPR14}
are comprised of short video clips capturing a single sports action,
making them unsuitable for our purpose.
Fortunately, the TRECVID MED 2011 \cite{MED11} dataset provides a large set of diverse videos collected directly from YouTube.
More importantly, these videos are not simple single clip videos; rather
they are complex events with rich interactions between
various sub-events within the same video \cite{Izadinia_ECCV12}.
Specifically, we learn our embeddings on the complete MED11 DEV and TEST sets comprised of
$40021$ videos. A subset of $256$ videos from the DEV and TEST set
was used for validation. The DEV and TEST sets are typical random assortments
of YouTube videos
with minimal constraints.

We compare our embeddings against different video representations 
for three video tasks:
video retrieval, complex event classification, and temporal order recovery.
All experiments are performed on the MED11 event kit
videos, which are completely disjoint from the training
and validation videos used for learning our embeddings.
The event kit is composed of $15$ event
classes with approximately $100-150$ videos per event, with a
total of $2071$ videos.

We stress that the embeddings are learned in a completely 
unsupervised setting and capture
the temporal and semantic structure of the data. We
do not tune them specifically to any event class and
$\sim 0.3\%$ of the DEV and TEST sets contain videos from each
category. This is not unreasonable, since any
large unlabeled video dataset is expected
to contain a small fraction of videos from every event.


%
%
%

\section{Video Retrieval}

In retrieval tasks, we are given a query, and the goal is to retrieve a set of related examples from a database.
We start by evaluating our embeddings on two types of retrieval tasks: event retrieval and temporal retrieval.
The retrieval tasks help to evaluate the ability of our embeddings to group together videos
belonging to the same semantic event class and frames that are temporally
coherent. 

\subsection{Event retrieval}
In the event retrieval scenario, we are given a query video from the MED11 event kit
and our goal is to retrieve videos that contain the same event from the remaining videos in the event kit.
For each video in the event kit, we sort all other videos in the dataset
based on their similarity to the query video using the cosine similarity metric, which we found to work best
for all representations.
We use Average Precision (AP) to measure the retrieval
performance of each video and provide
the mean Average Precision (mAP) over all videos in Tab.~\ref{tab:retrieval_results}.
For all methods, we uniformly sample
$4$ frames per video and represent the video as an average of
the features extracted from them. The different
baseline methods used for comparison are explained
below:

\vspace{2mm}

\begin{itemize}
    \vspace{-14pt}
  \item \emph{Two-stream pre-trained}: We use the two-stream CNN
    from \cite{Simonyan_2014} pre-trained on the UCF-101 dataset.
    The models were used to extract spatial and temporal features
    from the video with a temporal stack size of 5.
  \vspace{-14pt}
  \item \emph{fc6} and \emph{fc7}: Features extracted
    from the ReLU layers following the
    corresponding fully connected layers of a standard CNN
    model \cite{Krizhevsky_NIPS12} pre-trained on ImageNet.
  \vspace{-4pt}
  \item \emph{Our model (no temporal)}: Our model trained with no temporal context (Fig.~\ref{fig:system_figure_1}(c)).
  \vspace{-4pt}
  \item \emph{Our model (no future)}: Our model trained with no future context (Fig.~\ref{fig:system_figure_1}(b))
    but with multi-resolution sampling and hard negatives.
  \vspace{-4pt}
  \item \emph{Our model (no hard neg.)}: Our model trained without hard negatives
    from the same video.
  \vspace{-4pt}
  \item \emph{Our model}: Our full model trained with multi-resolution sampling
    and hard negatives.
  \vspace{-4pt}
\end{itemize}

\begin{table}
\begin{center}
\small
 \begin{tabular}{|l|c|}
  \hline
  Method &  mAP ( \%) \\
  \hline \hline
   Two-stream pre-trained \cite{Simonyan_2014} & 20.09\\
   fc6 &  20.08 \\ 
   fc7 &  21.24  \\ \hline \hline 
   Our model (no temporal) & 21.92 \\
   Our model (no future) & 21.30 \\ 
   Our model (no hard neg.) & 24.22\\ \hline \hline
  \textbf{Our model} & \textbf{25.07} \\ \hline 
  \end{tabular}
\end{center}
  \caption{Event retrieval results on the MED11 event kits.}
\label{tab:retrieval_results}
\end{table}

We observe that our full model outperforms other
representations for event retrieval.
We note that in contrast to most other representations trained on ImageNet,
our model is capable of being trained with large quantities of
unlabeled video which is easy to obtain.
This confirms our hypothesis that learning from
unlabeled video data can improve feature representations.
While the two-stream model also has the advantage of being trained specifically
on a video dataset, we observe that the learned representations do not
transfer favorably to the MED11 dataset in contrast to fc7 and fc6 features trained on ImageNet.
A similar observation was made in \cite{Xu_2015,Zha_2015},
where simple CNN features trained from ImageNet consistently
provided the best results.

Our embeddings capture the temporal regularities and patterns in videos
without the need for expensive labels,
which allows us to more effectively represent the semantic space of events.
The performance
gain of our full context model over
the representation without temporal
order shows the need
for utilizing the temporal information 
while learning the embeddings.

\begin{figure}[t!]
\centering
   \includegraphics[width=1\linewidth]{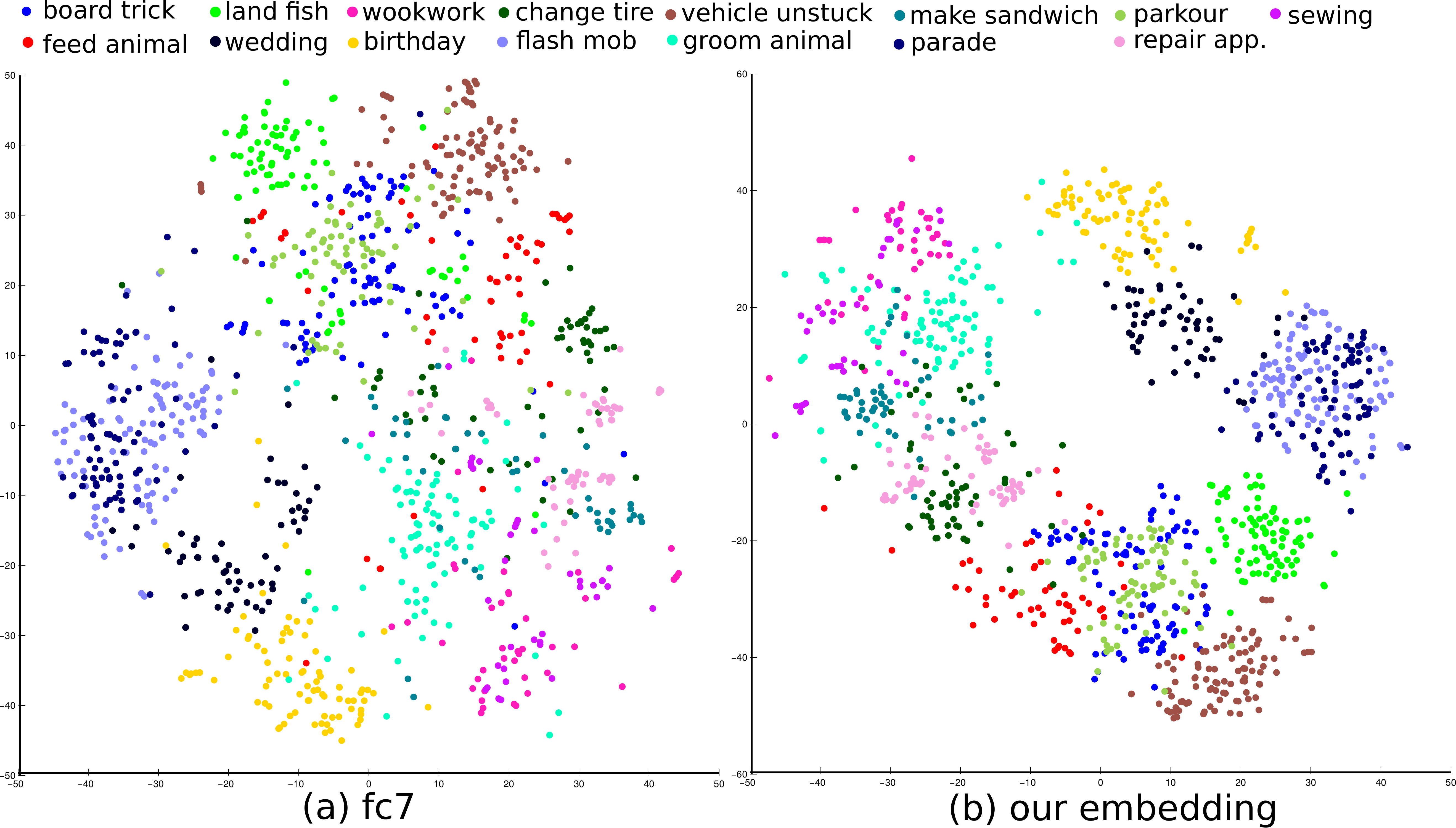}
   \caption{t-SNE plot of the semantic space for
(a) fc7 and (b) our embedding. The different colors correspond to 
   different events.}
\vspace{-8pt}
\label{fig:tsne_plots}
\end{figure}

\noindent \textbf{Visualizing the embedding space.}
To gain a better qualitative understanding of our learned embedding space,
we use t-SNE~\cite{Maaten_JMLR08} to visualize the embeddings in a 2D space.
In Fig.~\ref{fig:tsne_plots}, we visualize the fc7 features
and our embedded features by sampling a random set of videos from the event kits.
The different colors in the graph correspond
to each of the $15$ different event classes, as listed in the figure.
Visually, we can see that
certain event classes such as ``Grooming an animal", ``Changing a vehicle tire", and ``Making a sandwich" enjoy better clustering
in our embedded framework as opposed to the fc7 representation.

\begin{figure}[t!]
\centering
   \includegraphics[width=0.9\linewidth]{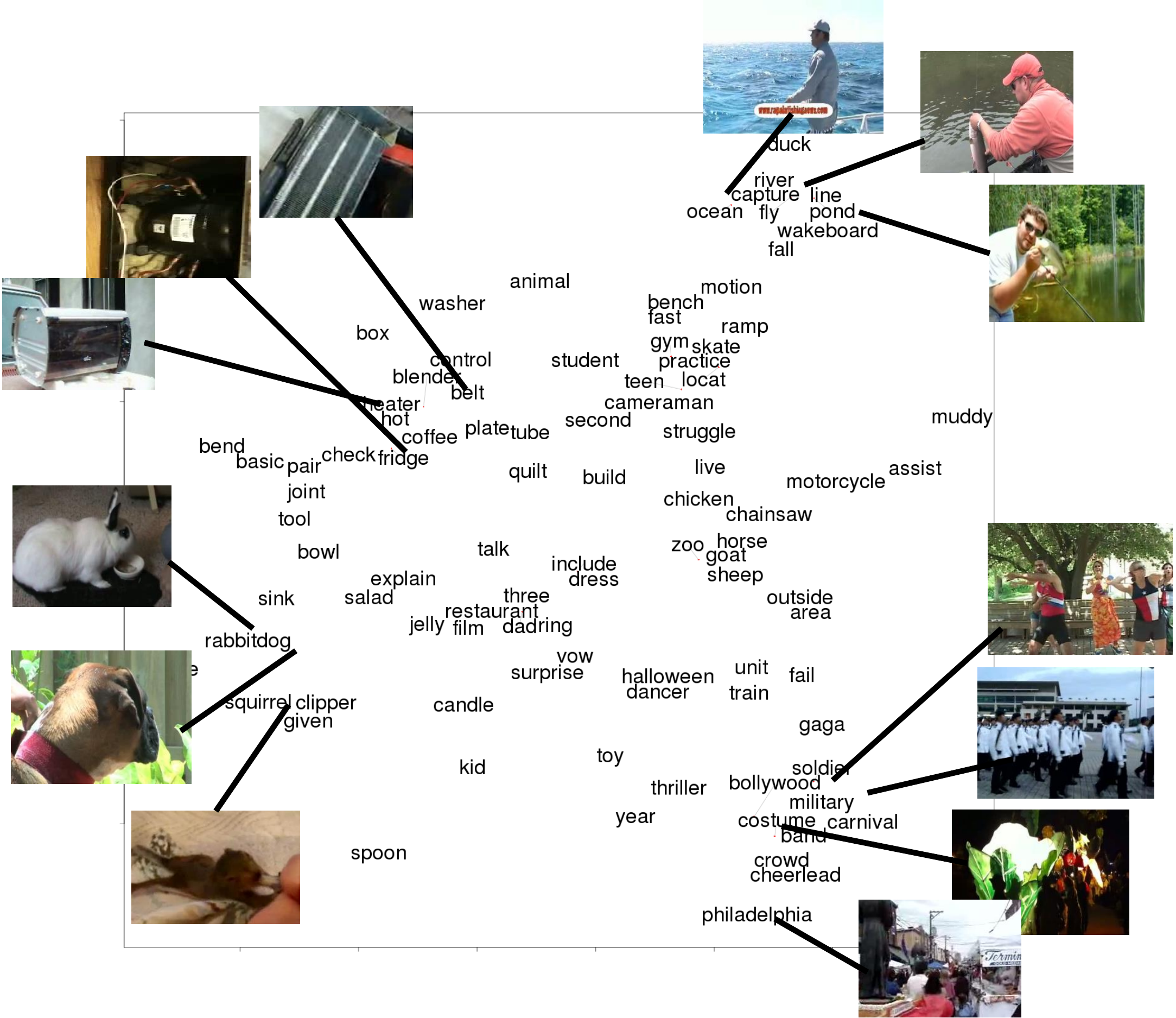}
   \caption{t-SNE visualization of words from synopses
     describing MED11 event kit videos.
   Each word
   is represented by the average of our embeddings corresponding
   to the videos associated with the word. We show sample video frames for a subset of the words.
 }
\label{fig:tsne_words}
\vspace{-8pt}
\end{figure}

Another way to visualize
this space is in terms of the
actual words used to describe the events.
Each video in the MED11 event
kits is associated with a short synopsis describing
the video. We represent each word from this synopsis collection by averaging
the embeddings of videos associated with that word.
The features are then used to produce a t-SNE plot as shown
in Fig.~\ref{fig:tsne_words}.
We avoid noisy clustering due to simple
co-occurrence of words by
only plotting words which
do not frequently co-occur in the same synopsis.
We observe many interesting patterns. 
For instance, objects such as ``river", ``pond"
and ``ocean" which provide the same context
for a ``fishing" event are clustered together.
Similarly crowded settings such as ``bollywood",
``military", and ``carnival" are clustered together.
This provides a visual clustering of the words
based on shared semantic temporal context.

\begin{figure}[t!]
\centering
   \includegraphics[width=1\linewidth]{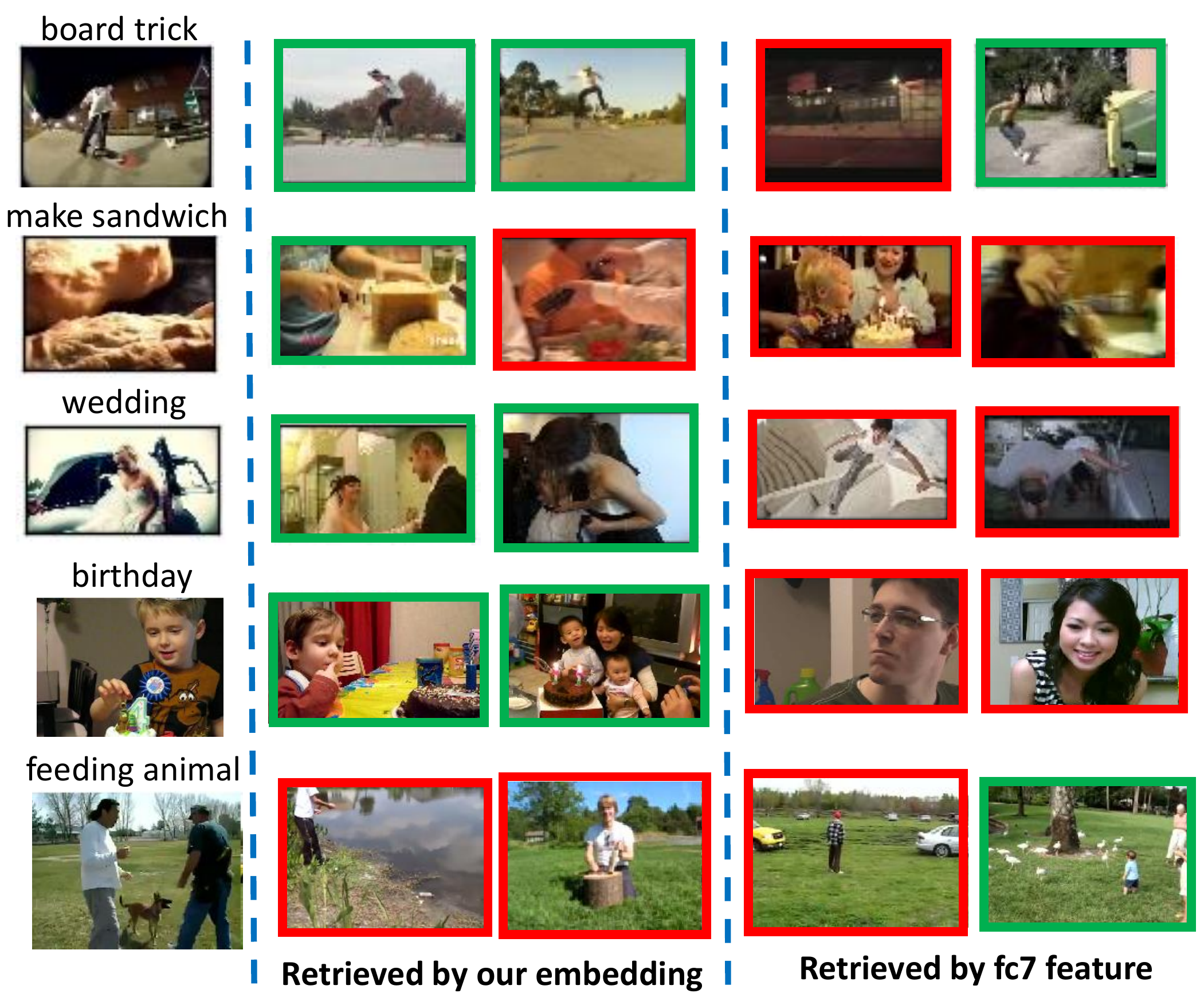}
   \caption{The retrieval results for fc7 (last two columns) and
     our embedding (middle two columns).
The first column shows the query frame and event,
while the top 2 frames retrieved from the remaining
videos are shown in the middle two column for our embedding,
and the last two columns for fc7.
The incorrect frames are highlighted in red, and correct frames in green. \vspace{-4mm}
   }
\label{fig:retrieval_results}
\end{figure}

\noindent \textbf{Event retrieval examples.}
We visualize the top frames retrieved for a few query frames
from the event kit videos in
Fig.~\ref{fig:retrieval_results}. The query frame is shown in the first
column along with the event class corresponding to the video.
The top 2 frames retrieved from other videos
by our embedding and by fc7 are shown in the first and
second columns for each query video, respectively.

We observe a few interesting examples where the query
appears visually distinct from the results retrieved by our
embedding. These can be explained by noting
that the retrieved actions might co-occur in the same context
as the query, which is captured by the
temporal context in our model. For instance, the frame of a ``bride near a car" retrieves frames of ``couple kissing".
Similarly, the frame of ``kneading dough" retrieves frames
of ``spreading butter".

\subsection{Temporal retrieval}
In the temporal retrieval task, we
test the ability of our embedding to capture the temporal
structure in videos. We sample four frames from different time instants
in a video and try to retrieve the frames in between the middle two frames.
This is an interesting
task which has potential for commercial applications
such as ad placements in video search engines.
For instance, the context at any time instant in a video
can be used to retrieve the most suited video ad
from a pool of video ads, to blend into the original video.

For this experiment, we use a subset of $1396$ videos from the
MED11 event kits which are at least $90$ seconds long.
From each video, we uniformly sample $4$ context frames,
$3$ positive frames from in between
the middle two context frames, and $12$ negative distractors from
the remaining segments of the video.
In addition to the $12$ negative distractors from the same video,
all frames from other videos are also treated as negative distractors.
For each video, given the $4$ context frames we evaluate our
ability to retrieve the $3$ positive frames from this large pool of distractors. 

We retrieve frames based on their cosine similarity
to the average of the features extracted from the
context frames, and use mean Average
Precision (mAP) as before.
We use the same baselines as the event retrieval
task. The results are shown in Tab.~\ref{tab:fillin_results}.

Our embedding representation which is trained
to capture temporal structure in videos is seen to
outperform the other representations.
This also shows their ability to capture
long-term interactions between events occurring
at different instants of a video.

\begin{figure*}[t!]
\centering
   \includegraphics[width=0.9\linewidth]{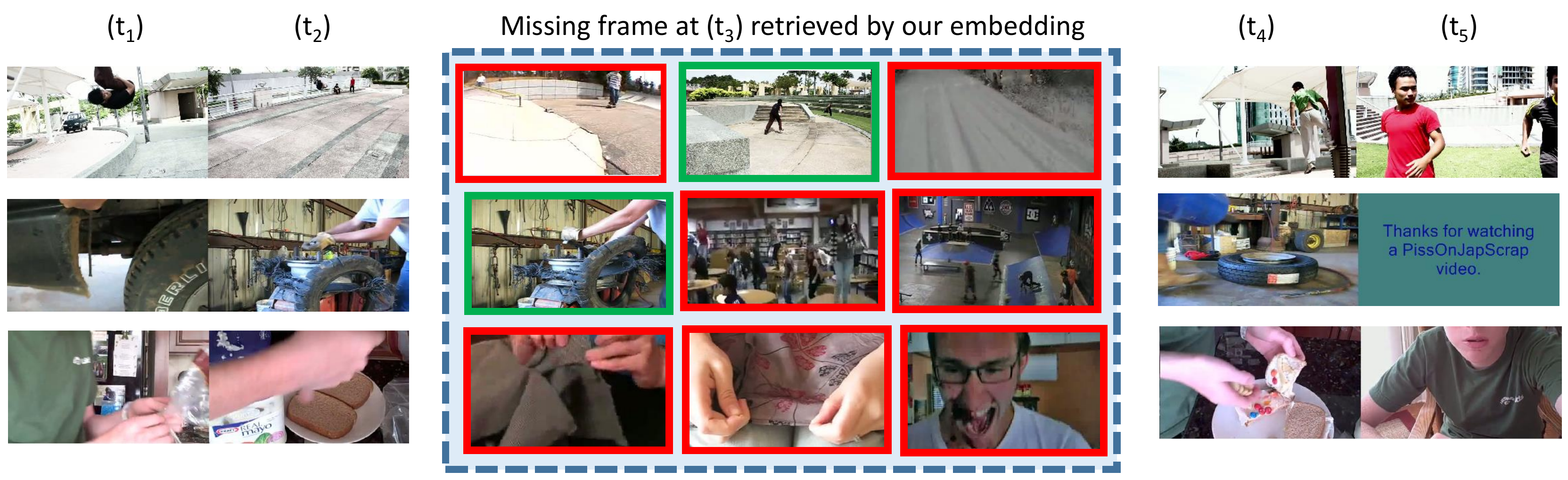}
   \caption{The retrieval results for our embedding model on the temporal retrieval task.
The first and last 2 columns show the $4$ context frames sampled from each video,
and the middle 3 columns show the top 3 frames retrieved by our embedding.
The correctly retrieved frames are highlighted in green, and incorrect frames highlighted in red. \vspace{-4mm}}
\label{fig:temporal_retrieval_results}
\end{figure*}

\noindent \textbf{Temporal retrieval examples.}
We visualize the top examples retrieved for a few temporal queries in
Fig.~\ref{fig:temporal_retrieval_results}. Here, we can see just how difficult this task is, as often
frames that seem to be viable options for temporal retrieval are not part of the ground truth.
For instance, in the ``sandwich" example, our embedding wrongly retrieves frames of human hands to keep up
with the temporal flow of the video. 

\begin{table}
\begin{center}
\small
 \begin{tabular}{|l|c|}
  \hline
  Method &  mAP ( \%) \\
  \hline \hline
   Two-stream pre-trained \cite{Simonyan_2014} & 20.11\\ 
   fc6 & 19.27 \\ 
   fc7 &  22.99 \\ \hline \hline
   Our model (no temporal) & 22.50 \\ 
   Our model (no future) & 21.71 \\ 
   Our model (no hard neg.) & 24.12 \\ \hline \hline
   \textbf{Our model}& \textbf{26.74} \\ \hline
  \end{tabular}
\end{center}
  \caption{Temporal retrieval results on the MED11 event kits.}
  \label{tab:fillin_results}
\vspace{-8pt}
\end{table}

\begin{table}[t!]
\begin{center}
\small
 \begin{tabular}{|l|c|}
  \hline
  Method &  mAP ( \%) \\
  \hline \hline
   Two-stream fine-tuned \cite{Simonyan_2014} & 62.99 \\ 
   ISA \cite{Le_CVPR11} & 55.87 \\ 
   Izadinia et al. \cite{Izadinia_ECCV12} linear & 62.63   \\ 
   Izadinia et al. \cite{Izadinia_ECCV12} full & 66.10 \\ 
   Raman. et al. \cite{Ramanathan_ICCV13} & 66.39 \\ 
   fc6 & 68.56 \\ 
   fc7 &  69.17 \\ \hline \hline
   Our model (no temporal) & 69.57 \\
   Our model (no future) &  69.22 \\ 
   Our model (no hard neg.) & 69.81 \\ \hline \hline
   \textbf{Our model}& \textbf{71.17} \\ \hline
  \end{tabular}
\end{center}
  \caption{Event classification results on the MED11 event kits.}
  \label{tab:classification_results}
\vspace{-12pt}
\end{table}

\section{Complex Event Classification}
The complex event classification task on the MED11 event kits is one
of the more challenging classification tasks.
We follow the protocol of \cite{Izadinia_ECCV12,Ramanathan_ICCV13} and
use the same train/test splits. Since the goal 
of our work is to evaluate the effectiveness of video frame
representations, we use a simple linear Support Vector Machine classifier for all methods.

Unlike retrieval settings, we are provided labeled training instances
in the event classification task.
Thus, we fine-tune the last two layers of
the two-stream model (pre-trained on UCF-101) on the
training split of the event kits, and found this to perform better than the pre-trained model.

In addition to baselines from previous tasks, we also compare with \cite{Izadinia_ECCV12}, \cite{Le_CVPR11} and \cite{Ramanathan_ICCV13}, with results shown in Tab.~\ref{tab:classification_results}.
Note that \cite{Izadinia_ECCV12,Ramanathan_ICCV13}
use a combination of multiple image and video features
including SIFT, MFCC, ISA, and HOG3D. Further, they also use
additional labels such as low-level events within each video.
In Tab.~\ref{tab:classification_results},
Izadinia et al. linear refers to the results
without low-level event labels.

We observe that our method outperforms ISA \cite{Le_CVPR11}, which is also a
unsupervised neural network feature representation.
Additionally, the CNN features trained from ImageNet seem to perform
better than most previous feature representations,
which is also consistent with the retrieval results and previous work \cite{Xu_2015,Zha_2015}.
Among the methods, the two-stream model holds the advantage of
being fine-tuned to the MED11 event kits.
However, our performance gain could be
attributed to the ability of our model to use
large amounts of unlabeled data to learn a
better representations.

\begin{figure*}[t!]
\centering
   \includegraphics[width=0.85\linewidth]{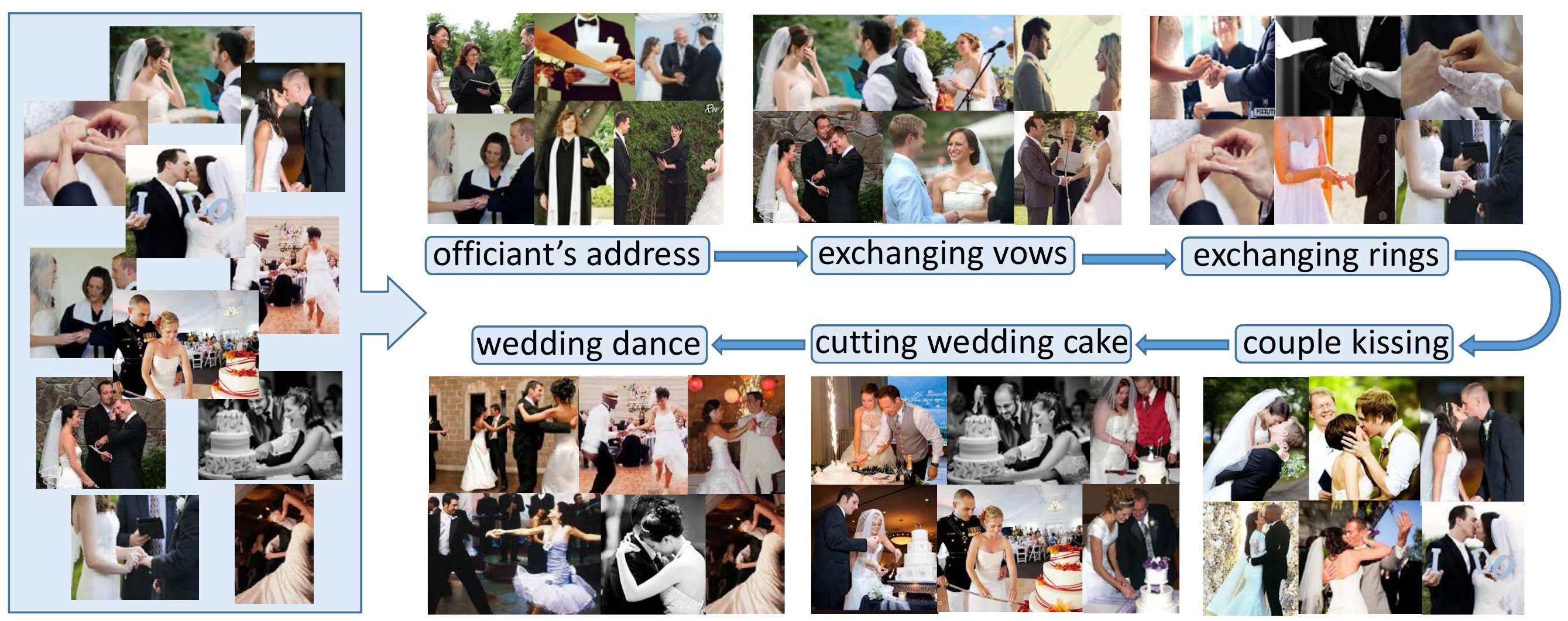}
   \caption{After querying the Internet for images of the ``wedding" event, we cluster them into sub-events and temporally organize the clusters using our model. On the left, we show sample images crawled for the ``wedding" event, and on the right the temporal order recovered by our model is visualized along with manual captions for the clusters.}
\label{fig:jumble_google_example}
\vspace{-8pt}
\end{figure*}

\section{Temporal Order Recovery}
An effective representation for video frames should be able to not only capture visual similarities,
but also preserve the structure between temporally coherent frames.
This facilitates holistic video understanding tasks beyond classification and retrieval.
With this in mind, we explore the video temporal order recovery task,
which seeks to show how the temporal interaction between different
parts of a complex event are inherently captured by our embedding.

\begin{figure}[t!]
\centering
   \includegraphics[width=1\linewidth]{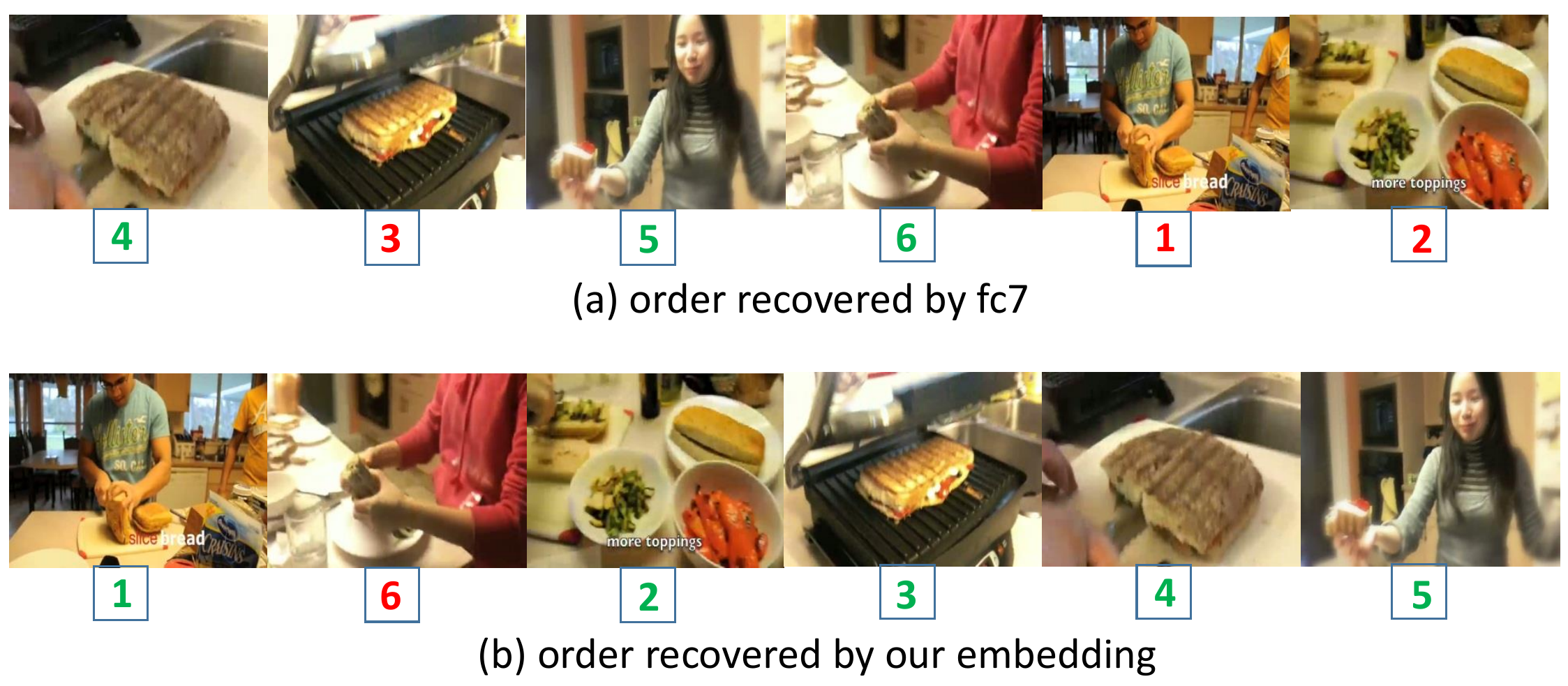}
   \caption{An example of the temporal ordering retrieved by fc7 and our method for
a ``Making a sandwich" video. The indexes of the frames
already in the correct temporal order are shown in green, and others in red.}
\label{fig:jumble_trecvid_example}
\vspace{-8pt}
\end{figure}

In this task,
we are given as input a jumbled sequence of frames
belonging to a video, and our goal is to order the frames into the correct sequence.
This has been previously explored in the context of
photostreams \cite{Kim_CVPR13}, and has potential for use
in applications such as album generation.


\begin{table}
\begin{center}
\small
 \begin{tabular}{|l|c|c|}
  \hline
  Method & 1.4k Videos  & 1k Videos\\
  \hline \hline
  Random chance & 50.00 & 50.00\\ 
  Two-stream \cite{Simonyan_2014} & 42.05 & 44.18\\
  fc6 & 42.43 & 43.33\\ 
  fc7 &  41.67 & 43.15\\ \hline \hline
  Our model (pairwise) & 42.03 & 43.72\\
  Our model (no future) & 40.91 & 42.98\\
  Our model (no hard neg.) & 41.02 & 41.95 \\ \hline \hline
  \textbf{Our model} & \textbf{40.41} & \textbf{41.13} \\ \hline
  \end{tabular}
\end{center}
  \caption{Video temporal order recovery results on the MED11 event
kits evaluated using the Kendell tau distance (normalized to 0-100).
Smaller distance indicates better performance.
The 1.4k Videos refers to the set of videos used in
the temporal retrieval task, and the 1k Videos refers to
a further subset with the most visually dissimilar frames.}
  \label{tab:video_jumble}
\vspace{-8pt}
\end{table}


\noindent \textbf{Solving the order recovery problem.}
Since our goal is to evaluate the effectiveness of various feature representations
for this task, we use a simple greedy technique to recover the temporal
order.
We assume that we are provided the first two frames in the video
and proceed to retrieve the next frame (third frame)
from all other frames in the video.
This is done by averaging the first two frames and retrieving the closest frame in cosine similarity.
We go on to greedily retrieve the fourth frame using the average of the second and third frames,
and continue until all frames are retrieved.
In order to enable easy comparison across all videos,
we sample the same number of frames ($12$) from each video
before scrambling them for the order recovery problem.
An example comparing our embeddings to fc7 is show in Fig.~\ref{fig:jumble_trecvid_example}.

\noindent \textbf{Evaluation.}
We evaluate the performance for solving the order recovery problem using the
Kendall tau \cite{Kendall_Bio38} distance between
the groundtruth sequence of frames and the sequence returned by the greedy method.
The Kendall tau distance is a metric that counts the number of
pairwise disagreements between two ranked lists; the larger the distance the more dissimilar the lists.
The performance of different features for this
task is shown in Tab.~\ref{tab:video_jumble}, where
the Kendall tau distance is normalized to be in the
range $0-100$.

Similar to the temporal retrieval setting, we use the subset
of $1396$ videos which are at least $90$ seconds long. These
results are reported in the first column of the table.
We observed that our performance was quite comparable
to that of fc7 features for videos with visually similar
frames like those from the ``parade" event, as they lack interesting
temporal structure. Hence, we also report results
on the subset of $1000$ videos which had the most
visually distinct frames. These results are shown
in the second column of the table.
We also evaluated the human performance of this task on a random subset of $100$ videos, and found the Kendell tau to be around $42$. This is on par with the performance of the automatic temporal order produced by our methods, and illustrates the difficulty of this task for humans as well.

We observe that our full context model trained with a temporal
objective achieves the best Kendall tau distance. This
improvement is more marked in the case of the 1k Videos
with more visually distinct frames. This shows
the ability of our model to bring together sequences of frames that should be temporally and semantically coherent.

\noindent \textbf{Ordering actions on the Internet.}
Image search on the Internet has improved to the point where we can find relevant images with
textual queries. Here, we wanted to investigate whether or not we could also temporally order images returned
from the Internet for textual queries that involve complex events. To do this,
we used query expansion on the ``wedding" query, and crawled Google for a large set of images.
Then, based on the queries, we clustered the images into sets of semantic clusters,
and for each cluster, averaged our embedding features to obtain a representation for the cluster.
With this representation, we then used our method to recover the temporal ordering
of these clusters of images. In Fig.~\ref{fig:jumble_google_example}, we show
the temporal ordering automatically recovered by our embedded features,
and some example images from each cluster.
Interestingly, the recovered order seems consistent with typical wedding scenarios.

\vspace{-6pt}
\section{Conclusion}

In this paper, we presented a model to embed video frames. We 
treated videos as sequences of frames and embedded them in a way which
captures the temporal context surrounding them.
Our embeddings were learned from a large
collection of more than $40000$ unlabeled videos, and have shown to
be more effective for multiple video tasks.
The learned embeddings performed better than other
video frame representations for all tasks.
The main thrust of our work is to push a framework for learning
frame-level representations from large sets of unlabeled video,
which can then be used for a wide range of generic
video tasks.

{\small
\bibliographystyle{ieee}
\bibliography{temporal_emb}
}

\end{document}